\documentclass[letterpaper, 10 pt, conference]{ieeeconf}  
\usepackage{amsmath,amsfonts,amssymb}
\usepackage{algorithmic}
\usepackage{algorithm}
\usepackage{array}
\usepackage[caption=false,font=normalsize,labelfont=sf,textfont=sf]{subfig}
\usepackage{textcomp}
\usepackage{stfloats}
\usepackage{url}
\usepackage{verbatim}
\usepackage{graphicx}
\usepackage{cite}
\usepackage{graphicx}
\usepackage{booktabs}
\usepackage{url}
\usepackage{balance}
\usepackage{color}
\def\eg{{\it e.g.}}

\def\ie{{\it i.e.}}

\def\vct#1{\mbox{\boldmath $#1$}}

\IEEEoverridecommandlockouts                              

\title{\LARGE \bf
Piggyback Camera: Easy-to-Deploy Visual Surveillance by\\Mobile Sensing on Commercial Robot Vacuums
}

\author{Ryo Yonetani$^{1}$
\thanks{$^{1}$Ryo Yonetani is with CyberAgent, Inc., Tokyo, Japan. 
        {\tt\small yonetani\_ryo@cyberagent.co.jp}}%
}

\begin{document}

\maketitle
\thispagestyle{empty}
\pagestyle{empty}

\begin{abstract}

This paper presents Piggyback Camera, an easy-to-deploy system for visual surveillance using commercial robot vacuums. Rather than requiring access to internal robot systems, our approach mounts a smartphone equipped with a camera and Inertial Measurement Unit (IMU) on the robot, making it applicable to any commercial robot without hardware modifications. The system estimates robot poses through neural inertial navigation and efficiently captures images at regular spatial intervals throughout the cleaning task. We develop a novel test-time data augmentation method called Rotation-Augmented Ensemble (RAE) to mitigate domain gaps in neural inertial navigation. A loop closure method that exploits robot cleaning patterns further refines these estimated poses. We demonstrate the system with an object mapping application that analyzes captured images to geo-localize objects in the environment. Experimental evaluation in retail environments shows that our approach achieves 0.83~m relative pose error for robot localization and 0.97~m positional error for object mapping of over 100 items.

\end{abstract}

\section{Introduction}

Robot vacuums represent one of the most successful commercial applications of robotics, yet their potential beyond autonomous cleaning remains largely unexplored. These robots have become ubiquitous in modern households and commercial spaces, with millions of units deployed worldwide. Leveraging this mobile robot infrastructure for visual surveillance applications, such as security~\cite{song2010surveillance,raty2010survey} or marketing~\cite{donepudi2020robots}, offers compelling advantages over installing extensive networks of static cameras. However, enabling such sensing capabilities faces significant challenges due to the proprietary nature of commercial robot vacuums. These systems are typically designed as closed platforms where accessing internal sensors, control algorithms, or navigation data requires substantial reverse engineering efforts and may compromise system reliability.

\begin{figure}[t]
    \centering
    \includegraphics[width=\linewidth]{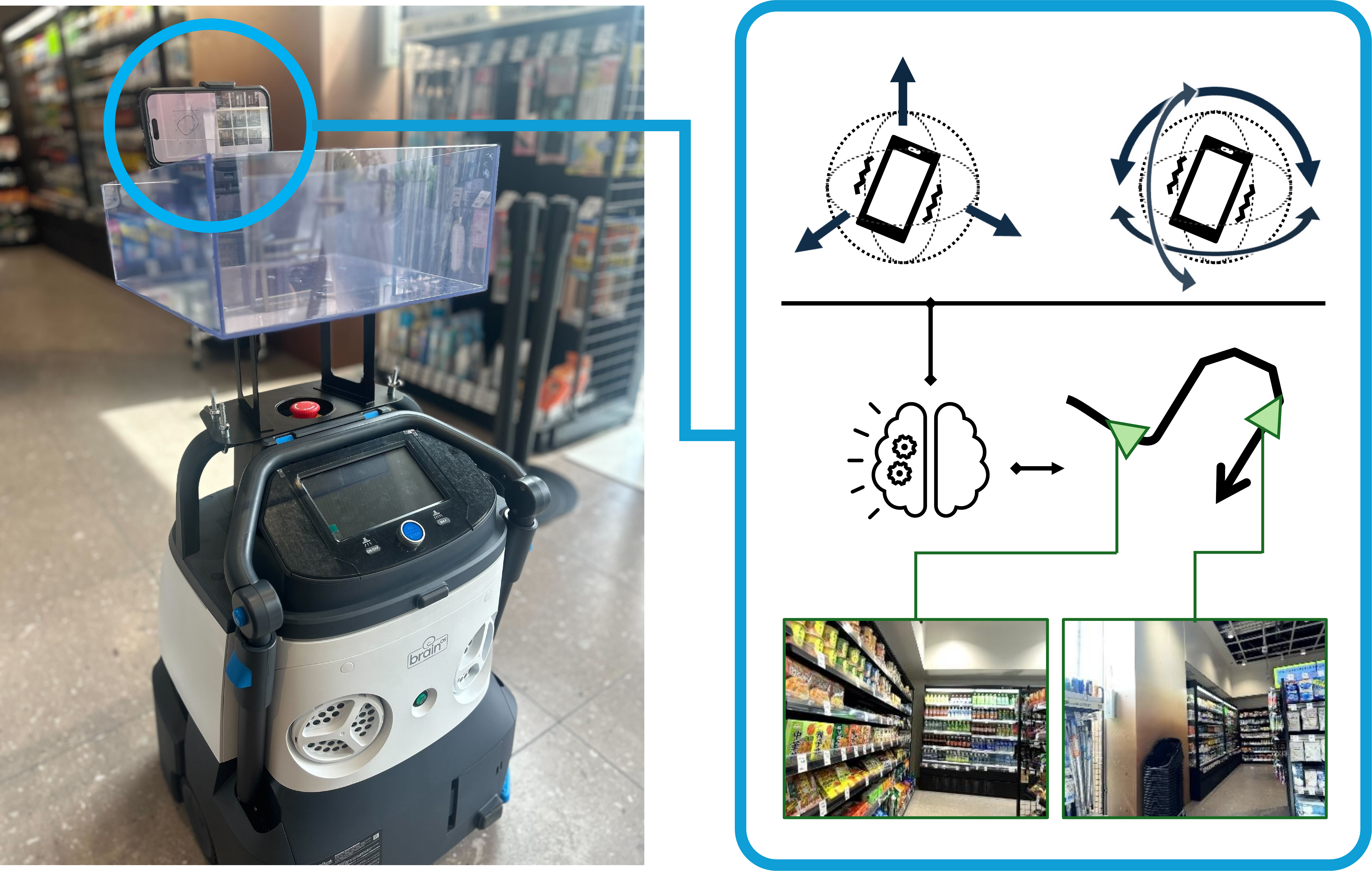}
    \caption{\textbf{Piggyback Camera System.} Left: A smartphone mounted on a commercial robot vacuum, enabling visual surveillance without hardware modifications. Right: Neural inertial navigation estimates the robot poses to capture images at regular spatial intervals.}
    \label{fig:teaser}
\end{figure}

We present \emph{Piggyback Camera}, a non-intrusive and easy-to-deploy system that mounts a smartphone on commercial robot vacuums to enable visual surveillance (\ie, the robot \emph{piggybacks} the smartphone, as shown in Fig.~\ref{fig:teaser}). Rather than modifying internal systems, our approach treats the robot vacuum as a black-box platform and adds sensing capabilities through the mounted smartphone. This design would make the system applicable to a variety of commercial robot vacuums regardless of manufacturer or model.

The primary technical challenge stems from estimating robot poses without access to internal state information. To associate images captured by the smartphone with their spatial locations and orientations, we must estimate the robot's poses throughout the environment. Crucially, this estimation must be computationally efficient given the limited resources of smartphone platforms. While Simultaneous Localization and Mapping (SLAM)~\cite{taketomi2017visual,wang2020approaches,zhan2020visual} methods provide robust pose estimation, they impose significant computational requirements. Similarly, continuous video recording for offline Structure from Motion (SfM)~\cite{saputra2018visual,schoenberger2016sfm} processing can severely impact smartphone storage and battery life, particularly in large environments.

\begin{figure*}
\centering
\vspace{1em}
\includegraphics[width=\linewidth]{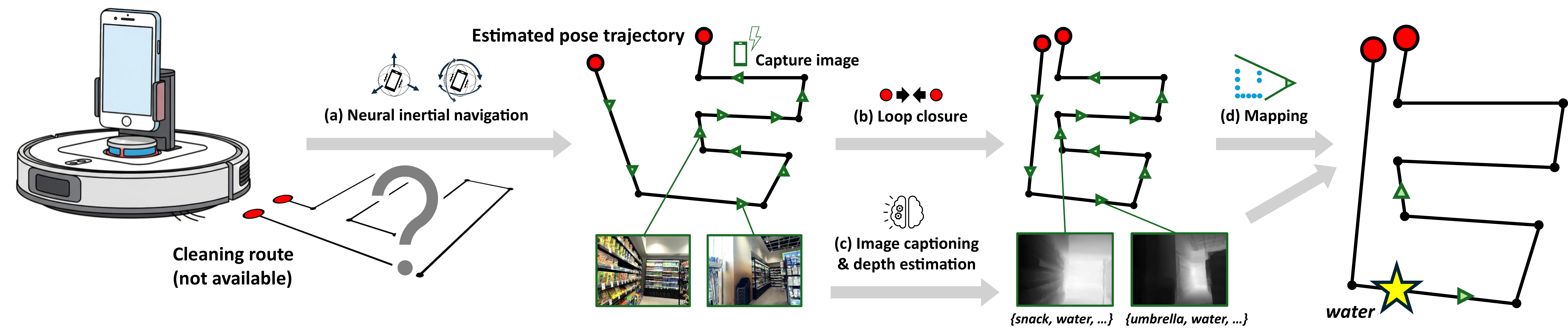}
\caption{\textbf{System Overview.} (a) Neural inertial navigation estimates the robot's pose trajectory using smartphone IMU data. (b) Loop closure refines the estimated poses by matching the start and goal positions. (c) The system captures images at regular spatial intervals and adopts image captioning and monocular depth estimation. (d) Integrating estimated poses and depth with the captions produces a geo-localized object map of the environment.}
\label{fig:overview}
\end{figure*}

Our system addresses these challenges by utilizing the smartphone's IMU sensors for efficient pose estimation via neural inertial navigation (Fig.~\ref{fig:overview}a). Similar to methods that estimate human locomotion from handheld devices~\cite{yan2020ronin,liu2020tlio,cao2022rio}, we estimate robot pose trajectories using the mounted smartphone's IMU data. To improve robustness against diverse and noisy IMU data, we propose \emph{Rotation-Augmented Ensemble (RAE)}, an efficient test-time data augmentation method that leverages the rotation equivalence property of neural inertial navigation~\cite{cao2022rio} to mitigate domain gap issues. We further develop a loop closure strategy that reduces estimation drift by constraining the robot's start and end positions to match at the charging station (Fig.~\ref{fig:overview}b). Using the estimated poses, the system captures images at regular spatial intervals (\eg, every 1~m and 90$^\circ$) to efficiently cover the environment. We demonstrate the approach through an object mapping application that uses multimodal large language models (LLMs) to identify objects in captured images (Fig.~\ref{fig:overview}c) and geo-localizes them within the environment (Fig.~\ref{fig:overview}d).

We evaluate the system through experiments with commercial robot vacuums in real retail environments. Results demonstrate that our approach achieves competitive positional accuracy with 0.83~m relative pose error compared to state-of-the-art visual localization methods~\cite{schoenberger2016sfm,pan2024global,wang2025vggt} and neural inertial navigation methods~\cite{yan2020ronin,cao2022rio}. For the object mapping application, the system achieves 0.97~m positional error when localizing over 100 items in the retail environment, outperforming baseline approaches.

\section{Related Work}
Our work relates primarily to camera pose estimation and neural inertial navigation. This section reviews existing approaches in these areas and highlights how our system addresses their limitations.

\subsection{Camera Pose Estimation}

Simultaneous Localization and Mapping (SLAM) and visual odometry techniques serve as the foundation for mobile robot localization (see~\cite{taketomi2017visual,wang2020approaches,zhan2020visual} for comprehensive surveys). Classical SLAM methods provide robust solutions for real-time pose estimation and mapping. Feature-based approaches~\cite{davison2007monoslam,mur2015orb,mur2017orb,campos2021orb} track sparse features across frames, while direct methods~\cite{engel2014lsd,engel2017direct} operate on pixel intensities without explicit feature extraction. Multi-modal systems such as RTAB-MAP~\cite{labbe2019rtab} extend these capabilities to incorporate both visual and LiDAR data for large-scale environments. Deep learning approaches have been increasingly applied to SLAM and visual odometry tasks~\cite{zhou2017unsupervised,li2018undeepvo,teed2021droid,yang2020d3vo,samsung2023dvi}. However, existing methods typically require significant computational resources and continuous visual tracking, which poses challenges for deployment on resource-constrained devices. These methods can also fail in environments with limited visual features or during rapid motion.

Structure from Motion (SfM) provides another approach to camera pose estimation~\cite{saputra2018visual}. Traditional methods like COLMAP~\cite{schoenberger2016sfm} and GLOMAP~\cite{pan2024global} reconstruct 3D scenes from unordered image collections through keypoint matching and bundle adjustment, while deep learning approaches have introduced end-to-end SfM pipelines~\cite{wei2020deepsfm,wang2023posediffusion,wang2024vggsfm}. More recently, 3D foundation models have emerged, beginning with DUSt3R~\cite{dust3r2024} for dense reconstruction, followed by MASt3R~\cite{mast3r2024}, VGGT~\cite{wang2025vggt}, and other methods~\cite{jang2025pow3r,wang20243d,yang2025fast3r} for enhanced camera pose estimation and scalability. However, SfM approaches are computationally intensive and typically require dense image coverage, making them unsuitable for mobile platforms with storage constraints.

Our approach addresses these limitations by leveraging neural inertial navigation for efficient real-time pose estimation and sparse image capture based on estimated poses.

\subsection{Neural Inertial Navigation}

\looseness=-1
Traditional inertial navigation systems estimate position through double integration of accelerometer and gyroscope measurements, but suffer from rapidly accumulating drift due to sensor noise and bias~\cite{woodman2007introduction}. Pedestrian Dead Reckoning (PDR) methods attempt to address this challenge through step detection and stride length estimation~\cite{harle2013survey}, but remain constrained by assumptions of periodic human walking patterns.  

Neural inertial navigation has emerged as a promising alternative. RoNIN~\cite{yan2020ronin} demonstrated that neural networks can learn complex motion patterns from IMU data without explicit sensor modeling, while TLIO~\cite{liu2020tlio} and IDOL~\cite{sun2021idol} integrate neural displacement estimates with extended Kalman filters to provide uncertainty quantification and improved robustness. Most existing methods focus primarily on human locomotion, while our Piggyback Camera system leverages neural inertial navigation to estimate camera poses for mobile robot applications. The most relevant work is RIO~\cite{cao2022rio}, which incorporated the rotation equivalence property of neural inertial navigation in a test-time training framework for domain adaptation. Our method, Rotation-Augmented Ensemble (RAE), also leverages this rotation equivalence property but operates without test-time training, enabling more efficient deployment on resource-constrained devices.

\section{Piggyback Camera System}
\subsection{System Overview}

\looseness=-1
The Piggyback Camera system consists of a smartphone mounted externally on a commercial robot vacuum. The smartphone serves as a sensing unit, utilizing its built-in IMU and camera to estimate robot poses and capture environmental images. This approach enables visual surveillance without requiring access to the robot's internal systems. We consider robot poses in the special Euclidean group $\textbf{SE}(2)$, consisting of 2D positions and 1D orientations (\ie, yaw), which is suitable for typical planar cleaning tasks performed by robot vacuums.

\looseness=-1
As illustrated in Fig.~\ref{fig:overview}, the system operates in two phases. During online operation while the robot performs its cleaning task, the system runs neural inertial navigation using the smartphone's IMU data to estimate the robot's pose trajectory and efficiently captures images at regular spatial intervals (Fig.~\ref{fig:overview}a). During offline analysis, the estimated positions are refined through loop closure that exploits the robot's predictable cleaning patterns (Fig.~\ref{fig:overview}b). The system also applies monocular depth estimation and image captioning to captured images (Fig.~\ref{fig:overview}c), which are combined with estimated poses to geo-localize objects found in the images (Fig.~\ref{fig:overview}d).

\subsection{Online Pose Estimation via Neural Inertial Navigation}
\label{sec:rae}
Online robot pose estimation employs neural inertial navigation~\cite{yan2020ronin}, which processes IMU data from the smartphone to estimate the robot's relative motion from its initial pose. The process begins by using raw accelerometer and gyroscope measurements to estimate device orientations. These measurements are then transformed into the Head-Agnostic Coordinate Frame (HACF), where the z-axis aligns with gravity and the x- and y-axes remain consistent throughout the recording. A pretrained neural network processes short-interval sequences of the HACF-transformed IMU data to estimate velocity for each interval. Finally, a Kalman filter estimates the robot's position using the velocity estimates and current position as observations.

The primary challenge in neural inertial navigation is velocity estimation error caused by domain gaps and inherent IMU noise across diverse robot motion patterns. Both factors contribute to significant drift in robot pose estimation. To address these issues, we propose a novel test-time data augmentation strategy called \emph{Rotation-Augmented Ensemble (RAE)} to mitigate domain gap problems online, and an offline pose refinement technique using loop closure to further reduce accumulated drift.

\subsection{Rotation-Augmented Ensemble (RAE)}
RAE leverages the rotation equivalence property of neural inertial navigation, inspired by \cite{cao2022rio}. When HACF-transformed IMU data is rotated by a certain angle around the z-axis (\ie, the gravity direction), the estimated velocity vector should be rotated accordingly. By rotating the velocity vector back to the original direction with the same angle, we obtain the velocity vector for the expected direction with augmented, diversified input. Given input IMU data, we perform this rotate-estimate-rotate-back operation for multiple angles and ensemble the results to obtain final velocity estimates. Unlike the test-time training strategy of \cite{cao2022rio}, RAE operates without fine-tuning the network online and is more computationally efficient.

Specifically, let $\vct{a}_t\in\mathbb{R}^3$ and $\vct{g}_t\in\mathbb{R}^3$ be acceleration and angular velocity at frame $t$, both represented in the HACF. A neural inertial navigation network $\mathcal{N}$ takes short-term sequences of these vectors $A_{t:t+\tau}=[\vct{a}_t,\dots,\vct{a}_{t+\tau}]$ and $G_{t:t+\tau}=[\vct{g}_t,\dots,\vct{g}_{t+\tau}]$ as input to produce a 2D velocity vector for the given interval: $\vct{v}_{t+\tau} = \mathcal{N}(A_{t:t+\tau}, G_{t:t+\tau})\in\mathbb{R}^2$. Let $Z(\vct{x}\mid\theta)$ denote the rotation operation of vector $\vct{x}$ by angle $\theta$ around the z-axis. We rotate the IMU data by angle $\theta_k$ as $Z(A_{t:t+\tau}\mid\theta_k) \triangleq [Z(\vct{a}_t\mid\theta_k), \dots, Z(\vct{a}_{t+\tau}\mid\theta_k)]$ and $Z(G_{t:t+\tau}\mid\theta_k) \triangleq [Z(\vct{g}_t\mid\theta_k), \dots, Z(\vct{g}_{t+\tau}\mid\theta_k)]$ and rotate the estimated velocity vector back to the original direction:
\begin{align}
    \vct{v}_{t:t+\tau}^{(k)} = Z\left(\mathcal{N}\left(Z(A_{t:t+\tau}\mid\theta_k), Z(G_{t:t+\tau}\mid\theta_k) \right)\mid-\theta_k\right).
\end{align}
RAE performs this rotation augmentation for $K$ angles $\{\theta_1,\dots,\theta_K\}$ uniformly sampled from $[-\pi, \pi)$ and takes the ensemble of the estimated velocity vectors:
\begin{align}
    \vct{v}_{t:t+\tau} = \phi\left(\left\{\vct{v}_{t:t+\tau}^{(1)},\dots,\vct{v}_{t:t+\tau}^{(K)}\right\}\right),
\end{align}
where $\phi$ is the ensemble function. We confirm that the simple median operation works robustly in practice.

\subsection{Offline Pose Refinement through Loop Closure}
\label{sec:loop_closure}
While RAE mitigates domain gap problems, inevitable localization drift from accumulated IMU noise still persists over time. We propose an offline method that refines estimated robot poses after the robot completes its cleaning task. The key insight is that commercial robot vacuums typically start and end their cleaning tasks at the same charge station. This allows us to perform loop closure by constraining the final estimated position to match the initial position, effectively reducing accumulated drift across the entire trajectory.

Let the estimated trajectory of length $T$ be $[\vct{p}_1,\dots,\vct{p}_T]$, where $\vct{p}_t\in\mathbb{R}^2$ is the estimated 2D position at frame $t$. We introduce correction parameters for each frame: $r_t \in [-\pi, \pi]$ represents the incremental rotation correction around the z-axis and $\vct{l}_t \in \mathbb{R}^2$ represents the 2D displacement correction. The refined positions are computed as:
\begin{align}
    \vct{p}'_t = \vct{p}_1 + \sum_{t'=2}^{t} Z(\vct{p}_{t'} - \vct{p}_{t'-1} \mid r_{t'}) + \vct{l}_{t}.
\end{align}

We optimize the following objective function with respect to the correction parameters $\{r_t, \vct{l}_t\}$:
\begin{align}
\mathcal{L}= \left\|\vct{p}'_T - \vct{p}_1\right\|^2 + \left\|\sum_{t=2}^T r_t\right\|^2 + \max_t \left|\vct{p}'_t - \vct{p}'_{t-1} - \vct{v}_t\right|,
\end{align}
where $\|\cdot\|$ denotes the L2 norm. This function consists of three terms: (1) loop closure constraint ensuring the final position $\vct{p}'_T$ matches the initial position $\vct{p}'_1$, (2) regularization term preventing excessive rotation, and (3) smoothness constraint preserving the estimated velocity. We implement this optimization using a small multi-layer perceptron (MLP) that predicts correction parameters from frame indices and optimize it via backpropagation similar to \cite{yonetani2024retailopt}.

\subsection{Object Mapping}
\label{sec:object_mapping}

\looseness=-1
Once the captured images are associated with refined poses, we geo-localize objects found in the images in the world coordinate system. As a concrete scenario, consider a retail application where shoppers seek to find items in stores. Being able to provide the locations of the items by their names can enhance the shopping experience. To achieve this application, we perform the following object mapping pipeline.

\textbf{Object Identification:} Our pipeline starts by identifying the names of items in images. We observe that standard object detection techniques are insufficient for this task, as their backbone models are typically trained to recognize generic object classes rather than specific item names. Instead, we utilize a multimodal LLM, such as OpenAI's GPT services, to extract the names of items from the images.

\textbf{Monocular Depth Estimation:} We utilize a monocular depth estimation network~\cite{yang2024depth} to estimate the metric depth map for each image. Given the focal length of the camera estimated beforehand, we convert the depth map to a 3D point cloud in the camera coordinate system. Since the camera poses are also estimated through neural inertial navigation, the 3D point cloud can be transformed to the world coordinate system.

\looseness=-1
\textbf{Object Mapping:} The two steps above provide us with a set of 3D points in the world coordinate system, each associated with item names extracted from the images. For each item, we cluster the 3D points associated with the same item name and compute their centroids as the geo-localized positions of the item. The final output is a set of item names and their corresponding positions in the world coordinate system.

\section{Experiments}
\label{sec:experiment}

We evaluated the Piggyback Camera system in a real-world retail environment, focusing on two main objectives: (1) assessing the effectiveness of the proposed neural inertial navigation method compared to state-of-the-art baselines on the pose estimation task and (2) demonstrating the practical utility of the system through the object mapping task.

\subsection{Data Collection}
We conducted experiments using a smartphone (iPhone 14 Pro) mounted on a commercial robot vacuum (SoftBank Robotics' Whiz i\footnote{\scriptsize{\url{https://www.softbankrobotics.com/jp/product/whiz/}}}) operating in a retail store of approximately 70~m$^2$. The smartphone was rigidly attached to the robot using a clamp-type mount, with the attachment point varying across recordings while maintaining a horizontal orientation and forward-facing camera direction. The robot operated at approximately 0.5~m/s to follow systematic coverage patterns throughout the store area.

During each cleaning session, we recorded IMU data at 50~Hz and video streams at 10~Hz from the smartphone. Ground-truth robot poses were obtained using visual-LiDAR SLAM~\cite{labbe2019rtab} running concurrently with data collection. Our dataset comprises 13 recordings (6.7~hours total) for training and 6 recordings (1.3~hours total) for evaluation.

\subsection{Implementation Details}
For neural inertial navigation (Sec.~\ref{sec:rae}), we trained the RoNIN-ResNet~\cite{yan2020ronin} model using both publicly available RoNIN data (23~hours) and our training data. The time interval $\tau$ was set to 64 frames (1.28~seconds). Training used the Adam optimizer~\cite{kingma2014adam} with the following hyperparameters: batch size of 32, learning rate of 0.0001, and 500 epochs. We adopted VQF~\cite{laidig2023vqf} for orientation estimation. For loop closure (Sec.~\ref{sec:loop_closure}), we implemented a three-layer MLP with 64 hidden units and ReLU activation, optimized for 100 epochs using Adam with a learning rate of 0.01. For object mapping (Sec.~\ref{sec:object_mapping}), we employed the Depth Anything V2 model~\cite{yang2024depth} fine-tuned for metric depth estimation\footnote{\scriptsize{\url{https://huggingface.co/depth-anything/Depth-Anything-V2-Metric-Indoor-Small-hf}}}, with maximum depth adjusted using training data to match the actual depth scale. Object identification was performed using OpenAI GPT-4.1\footnote{\scriptsize{\url{https://openai.com/index/gpt-4-1/}}} with the prompt: \emph{``Describe the names of all the products in the image while emphasizing texts if they exist.''} Camera focal length was obtained using COLMAP~\cite{schoenberger2016sfm} with calibration data collected offline.

\begin{figure*}[t]
    \vspace{1em}
    \centering
    \includegraphics[width=\linewidth]{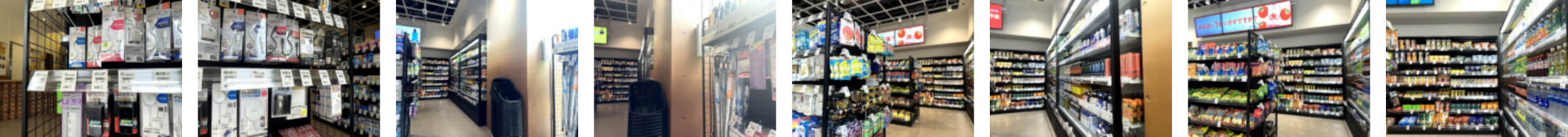}
    \caption{\textbf{Examples of Images.} Images sampled at regular spatial intervals from video streams taken by a smartphone during robot cleaning tasks.}
    \label{fig:example}
\end{figure*}

\subsection{Evaluation Setup}
\textbf{Sampling Images for Evaluation:} We estimated robot poses using the proposed RAE method with $K=3$ and extracted images from video streams in two ways to benchmark pose estimation performances: (1) every 0.5~m and 90$^\circ$ and (2) every 1.0~m and 90$^\circ$, following regular spatial intervals as described in the system overview. The average number of extracted images per recording was 251 and 112, respectively. Ground-truth poses for each image were obtained from the concurrent SLAM system. Fig.~\ref{fig:example} shows examples of images.

\textbf{Methods:} We evaluated the proposed method in four configurations: RAE and RAE with loop closure (RAE-LC), each with ensemble sizes $K$ set to 3 and 5. These methods were compared against visual localization baselines that directly estimated camera poses from the sampled images (COLMAP~\cite{schoenberger2016sfm}, GLOMAP~\cite{pan2024global}, and VGGT~\cite{wang2025vggt}) and neural inertial navigation baselines (RoNIN~\cite{yan2020ronin} and RIO~\cite{cao2022rio}).

\textbf{Pose Estimation Metrics:} Given a collection of sampled images with ground-truth poses, we evaluated pose estimation performances using Relative Translation Error (RTE) for 2D translation and Relative Rotation Error (RRE) for 1D rotation. After removing outliers, we aligned estimated poses to ground truth using optimal similarity transformation to minimize least-squares error between corresponding pose pairs. RTE is defined as the root mean squared error between aligned estimated poses and ground truth, while RRE is the average angle difference between aligned estimated and ground-truth orientations. For neural inertial navigation methods, we also computed RTE-metric, which removed the scale factor (\ie, fixed the scale factor to 1) from similarity transformation to assess metric accuracy. Finally, we evaluated coverage defined as the ratio of successfully estimated poses to total sampled images.

\textbf{Object Mapping Metrics:} For object mapping evaluation, we identified 112 items that are available in the unique location in the environment, extracted their names from image captioning results, and measured average positional errors of estimated item positions after pose alignment. Ground-truth item positions were obtained using the same object mapping pipeline but with SLAM pose trajectories.

\begin{table*}[t]
\centering
\caption{\textbf{Quantitative Results.} Relative Translation Error (RTE, RTE-metric), Relative Rotation Error (RRE), and coverage comparisons for different spatial sampling intervals.}
\label{tab:results}
\scalebox{1.0}{
\begin{tabular}{lcccccccc}
\toprule
 & \multicolumn{4}{c}{Grid: 0.5 m ($\sim 251$ images per recording)} & \multicolumn{4}{c}{Grid: 1.0 m ($\sim 112$ images per recording)}\\
\cmidrule(lr){2-5} \cmidrule(lr){6-9}
Method & RTE [m] & RTE-metric [m] & RRE [rad] & Coverage & RTE [m] & RTE-metric [m] & RRE [rad] & Coverage \\
\midrule
\multicolumn{8}{l}{\textbf{Visual Localization Baselines}} \\
COLMAP~\cite{schoenberger2016sfm} & 0.12 & N/A & 0.10 & 0.75 & 0.08 & N/A & 0.17 & 0.23 \\
GLOMAP~\cite{pan2024global} & 1.00 & N/A & 0.28 & 0.94 & 1.61 & N/A & 0.43 & 0.70 \\
VGGT~\cite{wang2025vggt} & \multicolumn{4}{c}{GPU Out of Memory} & 0.76 & N/A & 0.25 & 1.00 \\
\midrule
\multicolumn{8}{l}{\textbf{Neural Inertial Navigation Baselines}} \\
RoNIN~\cite{yan2020ronin} & 1.78 & 2.52 & 0.63 & 1.00 & 1.79 & 2.48 & 0.60 & 1.00 \\
RIO ($K=3$) & 1.40 & 1.51 & 0.52 & 1.00 & 1.43 & 1.52 & 0.51 & 0.99 \\
RIO ($K=5$) & 1.42 & 1.55 & 0.52 & 1.00 & 1.44 & 1.55 & 0.51 & 0.99 \\
\midrule
\multicolumn{8}{l}{\textbf{Proposed Methods}} \\
RAE ($K=3$) & 0.91 & 1.05 & 0.61 & 1.00 & 0.90 & 1.05 & 0.60 & 1.00 \\
RAE-LC ($K=3$) & 0.72 & 0.87 & 0.59 & 1.00 & 0.73 & 0.88 & 0.59 & 1.00 \\
RAE ($K=5$) & 0.91 & 1.08 & 0.62 & 1.00 & 0.91 & 1.08 & 0.60 & 1.00 \\
RAE-LC ($K=5$) & 0.66 & 0.82 & 0.58 & 1.00 & 0.66 & 0.83 & 0.57 & 1.00 \\
\bottomrule
\end{tabular}
}
\end{table*}

\begin{figure*}[t]
    \centering
    \includegraphics[width=\linewidth]{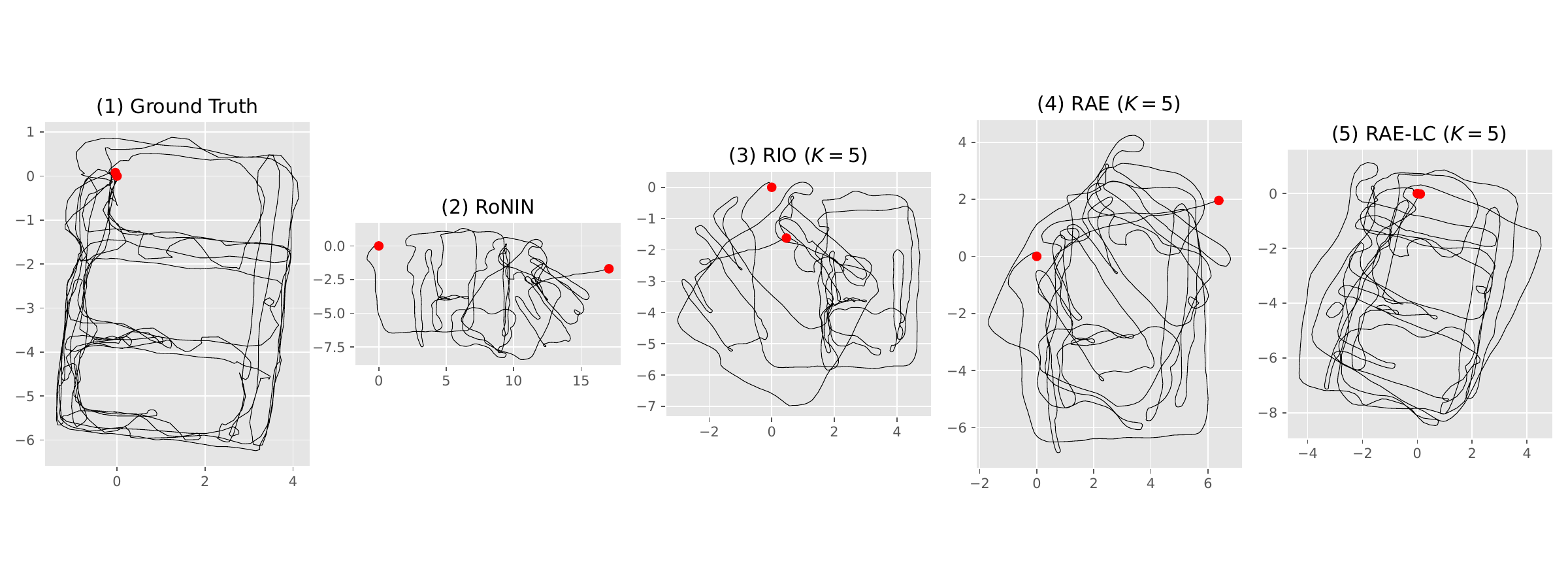}
    \caption{\textbf{Qualitative Results.} Trajectory estimation results in a retail environment. (1) Ground truth trajectory showing systematic cleaning pattern. (2,3) Baseline RoNIN and RIO methods exhibit significant drift. (4) RAE substantially reduces drift. (5) RAE-LC with loop closure achieves the best performance. Red dots indicate start/end positions. All trajectories are shown in original metric scale (meters).}
    \label{fig:results}
\end{figure*}

\begin{figure}
    \centering
\vspace{1em}
    \includegraphics[width=.8\linewidth]{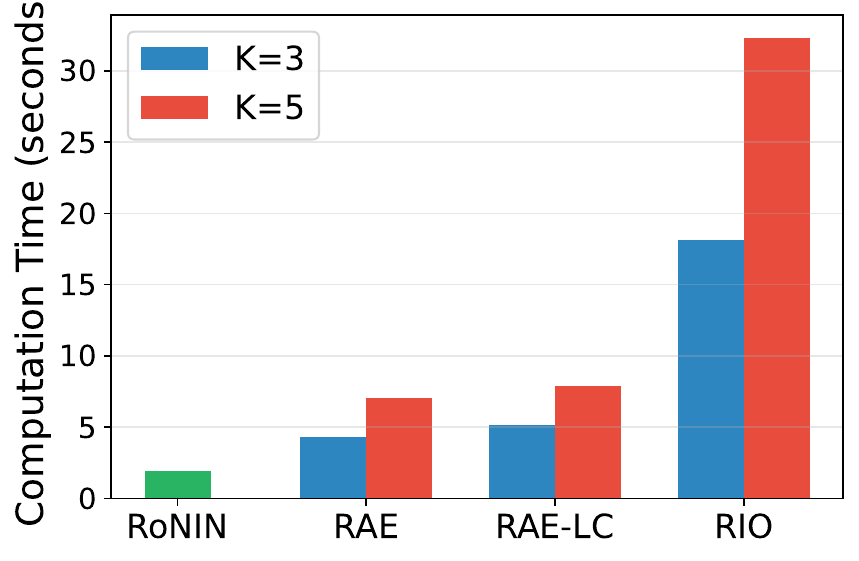}
    \caption{\textbf{Computation Time.} Computation time averaged over five trials to process 60~seconds (3000 frames at 50~Hz) of IMU data.}
    \label{fig:computation_time}
\end{figure}
\begin{table}[t]
    \centering
    \caption{\textbf{Object Mapping Results.} Mean and standard deviation of positional errors for 112 geo-localized objects.}
    \label{tab:object_localization_results}
    \begin{tabular}{lc}
        \toprule
        Method & Error (m) \\
        \midrule
        RoNIN~\cite{yan2020ronin} & 2.36 $\pm$ 1.57 \\
        RIO ($K=5$)~\cite{cao2022rio} & 1.55 $\pm$ 0.78 \\
        \midrule
        \textbf{RAE-LC ($K=5$)} & \textbf{0.97 $\pm$ 0.78} \\
        \bottomrule
    \end{tabular}
\end{table}

\subsection{Results}

Tab.~\ref{tab:results} summarizes the quantitative results comparing the proposed method against visual localization and neural inertial navigation baselines.

\textbf{Pose Estimation:} RAE-LC with $K=5$ achieved the best RTE-metric performance with 0.83~m error and outperformed neural inertial navigation baselines in RTE with 0.66~m error. For RRE, RIO ($K=5$) was the most accurate among neural inertial navigation methods, while RAE-LC ($K=5$) demonstrated comparable performance. Among visual localization baselines, COLMAP achieved the highest accuracy (0.08--0.12~m RTE and 0.10--0.17~rad RRE) but resulted in very low coverage unless images are collected sufficiently densely (0.23 coverage for 1.0~m grid sampling). GLOMAP and VGGT showed higher coverage but with trade-offs: GLOMAP performed poorly in RTE for 1.0~m grid sampling (1.61~m error), while VGGT could only process smaller image collections due to GPU memory constraints (confirmed with NVIDIA A100 80GB). Importantly, visual localization methods suffer from scale factor ambiguity and cannot provide metric accuracy, which is essential for object mapping.

\textbf{Ablation Study:} For the RAE, the offline loop closure refinement contributed to the overall improvement of RTE/RTE-metric. Increasing the ensemble size from $K=3$ to $K=5$ also provided lower translation errors. Note that the improvement by increasing $K$ nevertheless comes at the cost of increased computational overhead as shown below, indicating a trade-off between accuracy and efficiency.

\textbf{Computation Time:} Fig.~\ref{fig:computation_time} shows the computation time for neural inertial navigation methods to process 60 seconds of IMU data (3000 frames at 50~Hz), measured on CPU (Intel Core i7-1360P). RAE, RAE-LC, and RIO required more computation times compared to RoNIN due to the additional rotation augmentation. Nevertheless, RAE/RAE-LC are significantly faster than RIO as they do not perform online model fine-tuning. The overhead of loop closure refinement is negligible and less than one second.

\textbf{Qualitative Results:} Fig.~\ref{fig:results} visualizes trajectory estimation results. The ground truth trajectory demonstrates the systematic cleaning pattern typical of commercial robot vacuums, covering accessible store areas while avoiding obstacles. The RoNIN method exhibits significant drift and trajectory distortion, failing to maintain the proper spatial structure. The proposed RAE method substantially improves trajectory quality, better preserving the systematic cleaning pattern. With loop closure refinement, the final trajectory closely matches the ground truth, demonstrating effective drift correction.

\begin{figure*}[t]
    \centering
    \includegraphics[width=\linewidth]{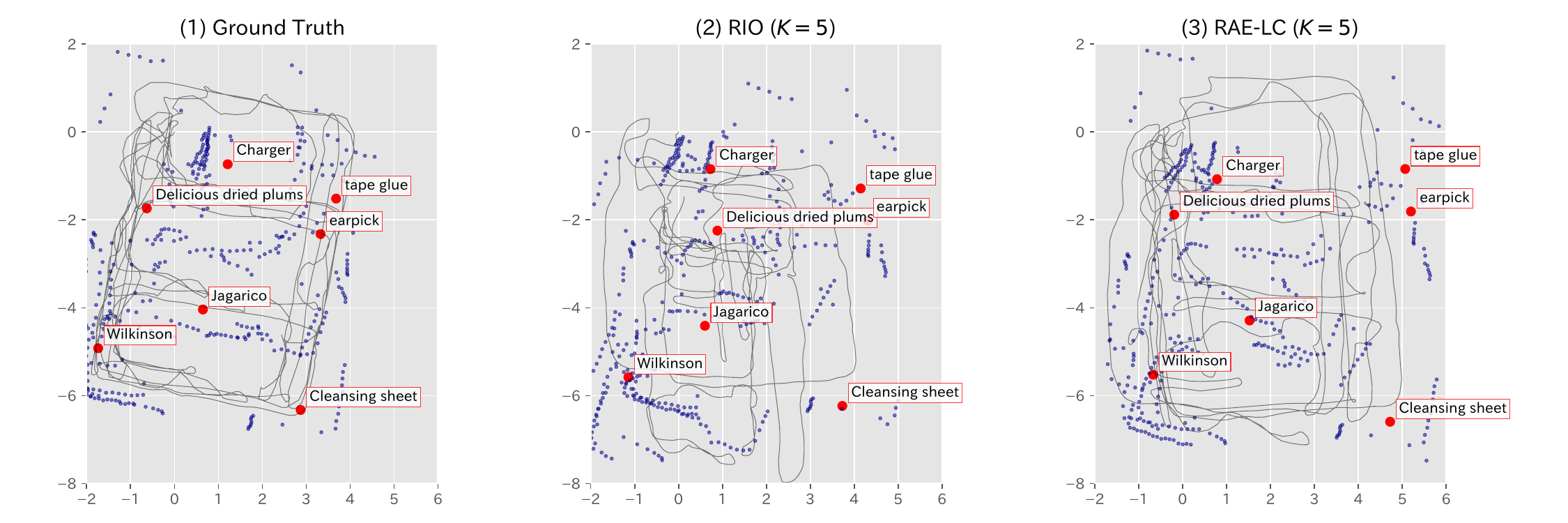}
    \caption{\textbf{Object Mapping Examples.} Geo-localized items (red dots) within estimated point clouds (blue small dots) around pose trajectories (gray lines).}
    \label{fig:object_localization_results}
\end{figure*}

\textbf{Object Mapping:} Tab.~\ref{tab:object_localization_results} summarizes the object mapping results, showing that RAE-LC ($K=5$) achieved the lowest positional error of 0.97~m, outperforming both RoNIN and RIO baselines. The qualitative results in Fig.~\ref{fig:object_localization_results} further illustrate the effectiveness of the proposed method in accurately localizing objects in the retail environment.

\subsection{Limitations}

The experimental results demonstrate the effectiveness of the proposed approach. However, several limitations suggest directions for future work.
\begin{itemize}
\item\textbf{State Space Constraints:} The current system is limited to planar $\textbf{SE}(2)$ pose estimation, which restricts applicability to flat surfaces. Extending to full $\textbf{SE}(3)$ estimation would enable deployment in more complex 3D environments. Moreover, the loop closure refinement relies on predictable robot behavior; more general techniques could enable adaptation to diverse robotic platforms.
\item\textbf{Object Mapping Limitations:} The object identification currently relies on multimodal LLMs, which may produce inconsistent or hallucinated results. A retrieval-augmented generation approach could improve object identification consistency by grounding LLM outputs in store-specific product catalogs and inventory databases.
\end{itemize}

\section{Conclusion}
This paper presented Piggyback Camera, an easy-to-deploy visual surveillance system that mounts smartphones on commercial robot vacuums. The system estimates robot poses using a rotation-augmented ensemble (RAE) strategy for robust neural inertial navigation and an offline loop closure method that exploits predictable cleaning patterns. Experimental evaluation demonstrated that the approach achieves 0.83~m relative pose error and enables accurate geo-localized object mapping with 0.97~m positional error in retail environments.

\balance
\bibliographystyle{IEEEtran}

\end{document}